\documentclass{article}
\usepackage{spconf,amsmath,graphicx}
\usepackage{cite}
\usepackage{amssymb,amsfonts}
\usepackage{algorithmic}
\usepackage{multirow}
\usepackage{booktabs} 
\usepackage{textcomp}
\usepackage{xcolor}
\usepackage{appendix}
\usepackage{url}
\usepackage{adjustbox}  
\usepackage{enumitem}

\setlist{leftmargin=4.0mm}

\makeatletter
\renewcommand\section{\@startsection {section}{1}{\z@}%
                                    {-2.6ex \@plus -0.5ex \@minus -1.4ex}%
                                    {1.0ex \@plus 0.5ex}%
                                    {\normalfont\Large\bfseries}}
\renewcommand\subsection{\@startsection{subsection}{2}{\z@}%
                                      {-1.6ex\@plus -0.25ex \@minus -.0ex}%
                                      {1.4ex \@plus 0.12ex}%
                                      {\normalfont\large\bfseries}}
\renewcommand\subsubsection{\@startsection{subsubsection}{3}{\z@}%
                                          {-0.25ex\@plus -0.1ex \@minus -.2ex}%
                                          {0.4ex \@plus .12ex}%
                                          {\normalfont\normalsize\bfseries}}
\renewcommand\paragraph{\@startsection{paragraph}{5}{\z@}%
                                      {0.75ex \@plus0.2ex \@minus1.01ex}%
                                      {-0.25em}%
                                      {\normalfont\normalsize\bfseries\slshape}}
\renewcommand\subparagraph{\@startsection{subparagraph}{6}{\parindent}%
                                         {0.5ex \@plus0.2ex \@minus .2ex}%
                                         {-0.2em}%
                                         {\normalfont\normalsize}}
\makeatother

\title{Large Language Model Based Generative Error Correction: \\
A Challenge and Baselines for
Speech Recognition, Speaker Tagging, and Emotion Recognition}

\name{
 \begin{tabular}{@{}c@{}}
 Chao-Han Huck Yang$^\dag$$^1$\thanks{$^\dag$Equal contribution.}, Taejin Park$^\dag$$^1$, Yuan Gong$^\dag$$^2$, Yuanchao Li$^\dag$$^3$, Zhehuai Chen$^1$, \\  Yen-Ting Lin$^4$, Chen Chen$^5$, Yuchen Hu$^5$, Kunal Dhawan$^1$, Piotr Żelasko$^1$, Chao Zhang$^6$, \\  Yun-Nung Chen$^4$, Yu Tsao$^7$, Jagadeesh Balam$^1$, Boris Ginsburg$^1$, Sabato Marco Siniscalchi$^8$, \\ Eng Siong Chng$^5$, Peter Bell$^3$, Catherine Lai$^3$, Shinji Watanabe$^9$, Andreas Stolcke$^{10}$\end{tabular}}

 \address{$^1$NVIDIA\quad$^2$MIT CSAIL\quad$^3$University of Edinburgh \\$^4$National Taiwan University\quad$^5$Nanyang Technological University\\ $^6$Tsinghua University\quad$^7$Academia Sinica\quad$^8$University of Palermo\quad$^9$CMU \quad$^{10}$Uniphore \\ 
 \small \textit{\{hucky,~taejinp\}@nvidia.com\quad yuangong@mit.edu\quad yuanchao.li@ed.ac.uk}\quad \textit{andreas.stolcke@uniphore.com}}

\begin{document}
%
\maketitle
\begin{abstract}
Given recent advances in generative AI technology, a key question is how large language models (LLMs) can enhance acoustic modeling tasks using text decoding results from a frozen, pretrained automatic speech recognition (ASR) model. To explore new capabilities in language modeling for speech processing, we introduce the generative speech transcription error correction (GenSEC) challenge. This challenge comprises three post-ASR language modeling tasks: (i) post-ASR transcription correction, (ii) speaker tagging, and (iii) emotion recognition. These tasks aim to emulate future LLM-based agents handling voice-based interfaces while remaining accessible to a broad audience by utilizing open pretrained language models or agent-based APIs. We also discuss insights from baseline evaluations, as well as lessons learned for designing future evaluations.


\end{abstract}

\section{Introduction}

Early statistical ASR systems based on the noisy channel model were conceived as utilizing two model components: acoustic model and language model (LM)~\cite{jelinek1976continuous}. First-pass decoding results could be subjected to postprocessing, or {\em rescoring}, to apply more powerful LMs or additional knowledge sources\cite{OstendorfEtAl:acl91}. With the introduction of end-to-end (E2E) ASR systems in the early 2020s, language modeling for post-ASR has become more complex, e.g., by also modeling the implicit {\em internal LM} of E2E ASR systems\cite{meng2021internal, ghodsi2020rnn, ma2024tuning, yu2023low}). With the advent of LLMs, however, post-ASR processing has become very attractive again, given the capacity of LLMs to model linguistic patterns, contextual influences, and even world knowledge as reflected in language.


More recently, LLMs and speech/language-model alignment methods have sparked considerable interest in new methods, such as cascaded LLM correction~\cite{yang2023generative}, for ASR and speech translation. LLM-based text-to-text generative error correction~\cite{yang2023generative, radhakrishnan2023whispering} has shown accuracy improvements over baseline rescoring methods, even surpassing $n$-best oracle performance, by bringing external knowledge to bear in ASR~\cite{chen2024hyporadise}, speech translation~\cite{hu2024gentranslate}, and image captioning~\cite{chan2023ic3, hirota2024descriptive}.

While a text-based ASR-LLM interface limits the richness of information utilized in post-processing, such as acoustic-prosodic expressions of speaker identity and paralinguistic properties, ASR outputs will still be sensitive to such information, especially when multiple hypotheses are output, effectively providing a text-based feature map that weakly reflects acoustic information \cite{yang2023generative}.

\begin{figure*}[ht!]
    \centering
    \includegraphics[width=0.7\linewidth]{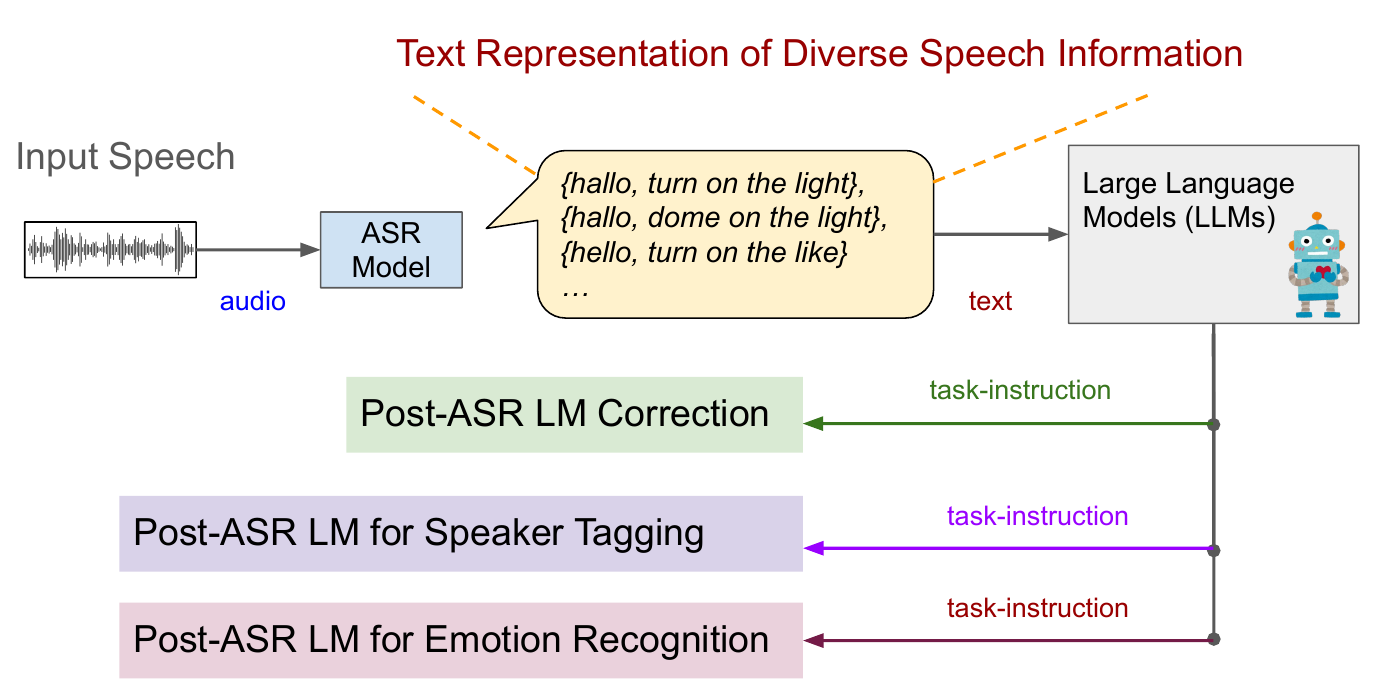}
    \caption{The framework for LLM postprocessing of ``text representation of speech'' via ASR-decoded information, for three tasks: speech recognition, speaker diarization, and emotion recognition.}
    \label{fig:overview}
    \vspace{-0.0cm}
\end{figure*}

Inspired by this observation, as well as by the early LLM-based studies cited, our challenge task aims to push research in two directions: (1) how large \textit{performance gains} could be achieved by applying cascaded ASR-LLMs, and (2) how well cascaded LLMs could perform \textit{tasks beyond word-level transcription}, such as recovering speaker and paralinguistic information. In other words, by leveraging LLMs, even text-only output from first-pass ASR system (such as available from a black-box API) might be enriched with paralinguistic and meta-information that is commonly thought to be encoded principally in acoustic-prosodic features.  Fig.~\ref{fig:overview} illustrates the three challenge tasks, highlighting the assumption that ASR hypotheses in textual form contain sufficient implicit acoustic information to perform these tasks.



The short-term goal of the challenge is to introduce new ASR-LM tasks to the speech community that leverage the latest developments in LLM-based post-ASR text modeling, potentially benefiting the design of voice-interface LLM agents that use only text-based encodings. Through this initiative, we aim to advance the understanding of LLM capabilities in implicit acoustic modeling within the spoken language processing community. In the longer term, our goal is to highlight the importance of audio and other modalities for speech processing and understanding, setting the stage for future challenges that go beyond the initial text-only framework.

By promoting ongoing cooperation and knowledge sharing, we seek to catalyze significant progress in the field of LLM-based voice interfaces, driving innovation and opening new avenues for research, development, and practical applications of ASR-LLM. This includes error type analysis and cross-lingual variants of these tasks. In the following sections we give more background and introduce the specifics of the challenge tasks.




\section{Background}
\subsection{Post-ASR Text Modeling}
This challenge considers an agent-based LLM application scenario with a fixed ASR interface. The focus is on characterizing how LLMs can enhance speech processing by leveraging the textual N-best hypotheses without explicit encoding of acoustic information. Additionally, we encourage participants to ``push the limits of language modeling for ASR'' given a setup based on a black-box ASR interface, which would be readily accessible through APIs, at modest cost.

The success of probabilistic language modeling systems in ASR can be traced back to several influential tools and frameworks, including SRILM~\cite{stolcke2002srilm}, CNTK~\cite{seide2016cntk}, and the LM components within Kaldi~\cite{povey2011kaldi}. These tools have significantly contributed to the development and enhancement of speech recognition technology by providing robust methods for handling the complexities of speech processing.
LLMs, on the other hand, have also benefited from democratized model inference (e.g., Claude) and open-source models (e.g., LLaMA) to establish end-to-end agent learning-based interfaces, such as AudioGPT~\cite{huang2024audiogpt}. For instance, work on task-activating prompting (TAP)~\cite{yang2023generative} illustrates that instruction-prompted LLMs for ASR can correct recognition errors by inferring phonetic confusions or grammatical variants from the ASR-decoded text.

To investigate this form of ASR-LLM pipeline, we introduce three tasks based on ASR-decoded text, which have been studied previously and were shown to benefit from combined acoustic and language modeling. In this challenge, participants can explore a training-free setup by optimizing instruction prompts for the speech tasks, or by hosting LLMs in their own compute environment. To examine the limits of the text-only modality for speech processing, we limit the first SLT challenge to text-to-text modeling without access to acoustic embeddings. The acoustic information will be accessed through ASR hypotheses ranked by acoustic confidence scores and error word-level or utterance-level error attributions. We hope this simple setup will entice researchers without a speech background to become active in speech processing through language modeling.

\subsection{Open Topics in LLM-based Speech Modeling}
To avoid test set data that may have leaked into the pretrained LLMs, we prepare a non-public test set for each challenge subtask. While LLMs hold promise for post-ASR correction, they are not without problems. One concern is the potential for introducing biases reflected in the training data, which could affect the accuracy and fairness of the corrected transcripts. Additionally, LLMs could produce enriched text that diverges from the intended meaning or introduces new types of errors, necessitating novel error analysis methodologies. These potential issues highlight the need for ongoing research and future challenges that assess the reliability and effectiveness of LLMs in ASR postprocessing.

Cross-modal setups will be incorporated into future versions of the challenge by providing acoustic embeddings~\cite{radhakrishnan2023whispering} or raw waveforms. By connecting the latest research and developments in speech language modeling with practical applications, the challenge promotes implementation, adoption, and understanding of cutting-edge language technologies, including in scarce-training or low-compute scenarios. Participants can expect to foster the development of innovative solutions at the intersection of speech and language technology.

\section{Challenge Description}
The GenSEC challenge at IEEE SLT 2024 consists of three tasks for post-ASR language modeling:
\begin{itemize}
\setlength\itemsep{0.05em}
    \item \textbf{Task 1: Post-ASR Output Correction by LLM} 

    The goal of this task is to map from N-best \textbf{H}ypotheses to ground \textbf{T}ruth transcriptions (H2T), similar to the setup in Yang et al.~\cite{yang2023generative}. The training set includes recognition scores from various pretrained end-to-end ASR models and N-best hypotheses. Participants are allowed to use N-best hypotheses and their scores for re-ranking or generative correction to produce final transcriptions. 
\end{itemize}

\begin{itemize}
    \item \textbf{Task 2: Post-ASR Speaker Tagging Correction}
    
This task aims at correcting the speaker tags in the output of a speaker-attributed (multi-speaker) ASR system. Speaker tagging in Task 2 refers to the speaker indices or anonymized speaker names (e.g., ``speaker-A'', ``speaker-2'') used to identify who spoke which words. We will provide errorful speaker-attributed transcripts produced by a multi-speaker ASR system. Participants in Task 2 are asked to submit corrected versions of the transcripts with accurate speaker tagging. A metric that gauges both speaker tagging and ASR accuracy will be used for evaluation. Similar to the other tasks, the current version of the Track-2 challenge allows use of the text modality only. 

    

\end{itemize}

\begin{itemize}
    \item \textbf{Task 3: Post-ASR Speech Emotion Recognition}
    
    This task aims to achieve utterance-level speech emotion recognition (SER) based on errorful ASR transcripts. Participants will develop ASR error correction methods combined with traditional deep-learning-based SER models, design novel prompt templates utilizing LLMs for SER, or utilize any other methods based on text input. Participants are encouraged to use the conversation as context to predict the emotion of a target utterance. 
 
\end{itemize}
\begin{figure}
    \centering
    \includegraphics[width=0.80\linewidth]{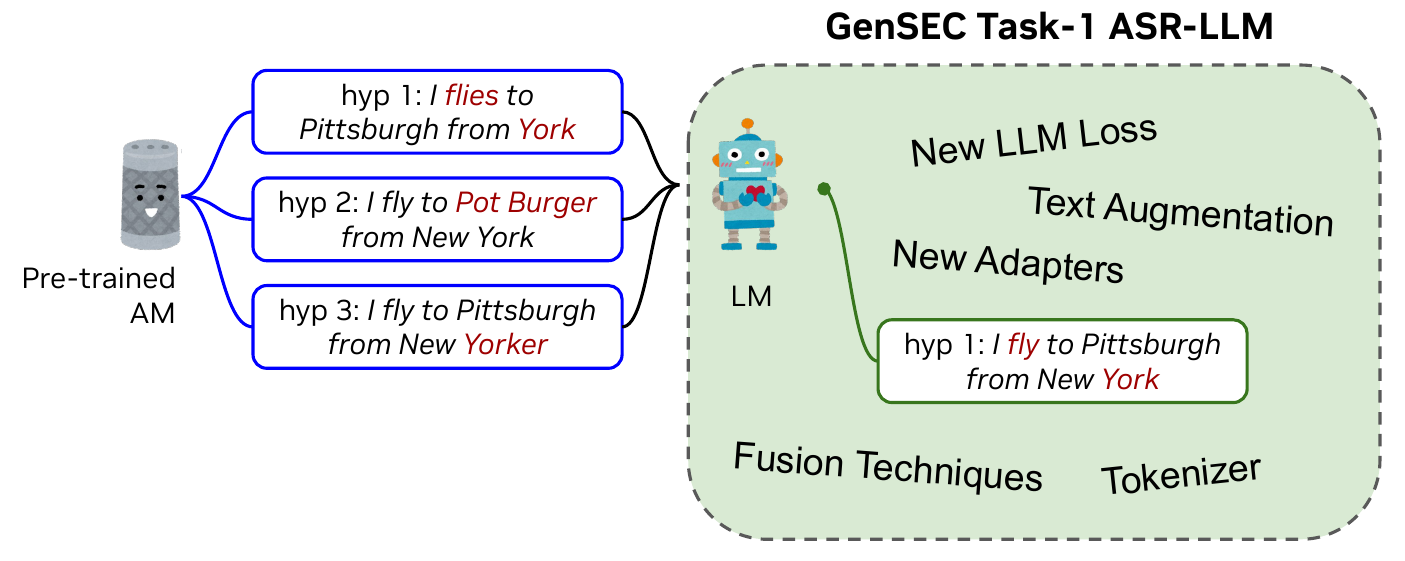}
    \caption{Example Task 1 approach: post-speech recognition error correction with different techniques on LLMs.}
    \label{fig:pipeline}
    \vspace{-0.0cm}
\end{figure}

\section{Task Description}
\subsection{Task 1 on LLM for Post-ASR Output Correction}

\textbf{Background:} Language model (LM) rescoring has been employed widely and for a variety of ASR technologies to improve ASR results, usually achieving good performance gains  \cite{OstendorfEtAl:acl91,stolcke2006recent,arisoy2015bidirectional,xiong2017toward}. In this approach, an external LM is trained separately and used to re-score the N-best hypotheses generated by the ASR system. While text error correction (TEC) has been explored \cite{dahlmeier2012better,yang2023generative}, ASR error correction is distinct due to the variability and distinct patterns of spoken language \cite{aksenova2021might}. Neural models have been used widely with E2E models for text error correction or normalization \cite{zhang2019neural, guo2019spelling, ma2023n}. These models often use beam search to generate new estimates, and can usually handle text normalization and denormalization of spelling errors.

\textbf{Motivation:} As shown in Fig.~\ref{fig:pipeline}, with Task 1 we aim to explore the limits of ASR-LLM error correction, as well as how best to utilize the ambiguity conveyed by N-best output. 


\textbf{N-best dataset:} The N-best open source corpus HyPoradise~\cite{chen2024hyporadise} will be made open-source under the MIT license. This includes HyPoradise training sets ($316.8$k pairs), development sets such as Librispeech-test-clean ($2.6$k pairs) and WSJ-dev93 ($503$ pairs), and evaluation sets including Librispeech-test-other ($2.9k$ pairs) and WSJ-dev93 ($333$ pairs).

\textbf{Baseline:} We provide pretrained 1st-pass and 2nd-pass models. The details of existing engineering pipelines are listed below. Training code has been released\footnote{\url{https://github.com/Hypotheses-Paradise/Hypo2Trans}} and the pretrained LLaMA2-7B model\footnote{\url{https://huggingface.co/GenSEC-LLM}.}
 has been released.

\textbf{Evaluation:} The challenge participants are allowed to apply their own 2nd-pass model to the provided ASR hypotheses decoded by beam search using Whisper.

\begin{table}[ht]
\centering
\caption{Task-1 WER ($\%$) of post-ASR LM correction on the HyPoradise~\cite{chen2024hyporadise} dataset.}
\begin{tabular}{rcc}
\toprule
           & \textbf{train}      & \textbf{test}     \\ \midrule
Whisper-1.5B (first-pass) w/o LM   & 10.43 & 11.82 
\\ \midrule 
$N$-best Oracle   & 9.61 & 9.32 \\ \midrule\midrule
Reranking LM: T5-750M   & 9.90 & 9.74 \\ \midrule
Correction LM: T5-750M   & 9.21 & 9.05 \\ \midrule
Correction LM: LLaMA-13B & 8.62 & 8.63 \\ \midrule
Correction LM: LLaMaA2-7B & 8.71 & 8.33 \\ \bottomrule
\label{tab:task1_results}
\end{tabular}
\end{table}


The WER of the corrected hypotheses is used for evaluation. This WER is compared to two
``oracle'' WERs calculated from the N-best inputs, namely, 1) the lowest WER achievable by picking the best hypothesis from each N-best list, and 2) the compositional oracle method ocp: the achievable WER using ``all tokens'' in the N-best hypothesis list. The former can be viewed as a lower bound on re-ranking methods, while the latter denotes the lower bound using elements already occurring in the list. To understand the effect of text normalization (punctuation and capitalization, P\&C) on ASR performance, both normalized and unnormalized (P\&C) WERs are reported. 


\begin{table}[t]
\caption{Task-1 dataset statistics: number of hypothesis-transcription pairs and average utterance length.
}
\centering
\resizebox{1.0\columnwidth}{!}{
\begin{tabular}{cc|ccc|ccc}
\toprule[1.5pt]
\multicolumn{2}{c|}{Domain} & \multirow{2}{*}{Training Set}& \multirow{2}{*}{\# Pairs}&\multirow{2}{*}{Length} & \multirow{2}{*}{Test Set} & \multirow{2}{*}{\# Pairs} &\multirow{2}{*}{Length}\\
Source & Category &  &  &  & & & \\
\midrule[1.5pt]
\multirow{2}{*}{LibriSpeech} & \multirow{2}{*}{Audiobooks} & \multirow{2}{*}{\emph{train-960}} & \multirow{2}{*}{88,200} & \multirow{2}{*}{33.7} & \emph{test-clean} & 2,620 & 20.1 \\
&&&&& \emph{test-other} & 2,939 & 17.8 \\
\midrule
CHiME4 & Noise & \emph{train} & 8,738 & 17.0 & \emph{test-real} & 1,320 & 16.4 \\
\midrule
\multirow{2}{*}{WSJ} & \multirow{2}{*}{Business news} & \multirow{2}{*}{\emph{train-si284}} & \multirow{2}{*}{37,514} & \multirow{2}{*}{17.5} & \emph{dev93} & 503 & 16.7 \\
&&&&& \emph{eval92} & 333 & 17.3 \\
\midrule
SwitchBoard & Telephone & \emph{train} & 36,539 & 11.8 & \emph{eval2000} & 2,000 & 11.8\\
\midrule
CommonVoice & Accented English & \emph{train-accent} & 49,758 & 10.5 & \emph{test-accent} & 2,000 & 10.5 \\
\midrule
Tedlium-3 & TED talk & \emph{train} & 47,500 & 12.6 & \emph{test} & 2,500 & 12.6\\
\midrule
LRS2 & BBC audio & \emph{train} & 42,940 & 7.6 & \emph{test} & 2,259 & 7.6 \\
\midrule
ATIS & Airline info. & \emph{train} & 3,964 & 12.4 & \emph{test} & 809  & 11.3 \\\midrule
CORAAL & Interview & \emph{train} & 1,728 & 24.2 & \emph{test} & 100 & 24.0  \\
\midrule[1.5pt]
\multicolumn{2}{c|}{Total} & \emph{train} & 316,881 & 18.1 & \emph{test}& 17,383 & 14.1 \\
\bottomrule[1.5pt]
\end{tabular}}
\label{statistics}
\end{table}

\subsection{Task 2: Post-ASR Speaker Tagging Correction}

\textbf{Background:} While the use of lexical cues in speaker diarization, speaker turn detection, and speaker segmentation has been explored previously, it is still less commonly used than acoustic-only speaker diarization. Early studies in this area, such as those presented in \cite{canseco2004speaker, canseco2005comparative}, utilized linguistic patterns to identify speakers during the diarization process.  

Several studies have enhanced speaker segmentation and clustering accuracy by integrating ASR output to leverage lexical cues \cite{park2018multimodal, xia2022turn, khare2022asr}. Furthermore,  lexical cues can be incorporated into speaker diarization by combining speaker turn probabilities based on both audio and text during the clustering phase \cite{park2019speaker}. Alternatively, spoken words and speaker channels can be decoded jointly, thus utilizing lexical cues implicitly for diarization \cite{shafey2019joint, kanda2022transcribe}.  

More recently, the study presented in \cite{cheng2023exploring} introduced semantic information through neural embeddings generated by a spoken language processing (SLP) unit. Subsequently, a multimodal (audio-text) speaker change detector was proposed \cite{jung2023encoder}, along with a speaker error correction (SEC) system \cite{paturi2023lexical} based on a pretrained language model.

Due to the recent popularity of LLMs, the multi-speaker ASR and speaker diarization community has also begun employing LLMs to enhance performance. One framework established for this purpose fine-tunes PaLM 2-S \cite{chowdhery2023palm} to correct speaker diarization errors from GCP’s Universal Speech Model \cite{zhang2023google}, which uses Turn-to-Diarize \cite{xia2022turn} for speaker diarization \cite{wang2024diarizationlm}. More recently, an ensemble of LLMs has been proposed to correct speaker diarization outputs \cite{efstathiadis2024llm}.

\textbf{Motivation:}  
As discussed for Task 1, LM rescoring for ASR has been widely studied and adopted, as external language models can be trained on relatively larger text-only datasets. Numerous studies have demonstrated the benefits of using LLMs for speaker diarization correction. Despite an abundance of research, there has been no standardized evaluation of multi-speaker error correction systems. We believe that our GenSEC Challenge Task 2 is timely in filling this gap.  To lower the bar for entry, we focus on the text modality, excluding acoustic or visual modalities in this first round. Therefore, in Task 2, we employ the system proposed in \cite{park2023enhancing} without utilizing acoustic information from the (acoustic-only) speaker diarization system.

\begin{figure*}
    \centering
    \includegraphics[width=0.85\linewidth]{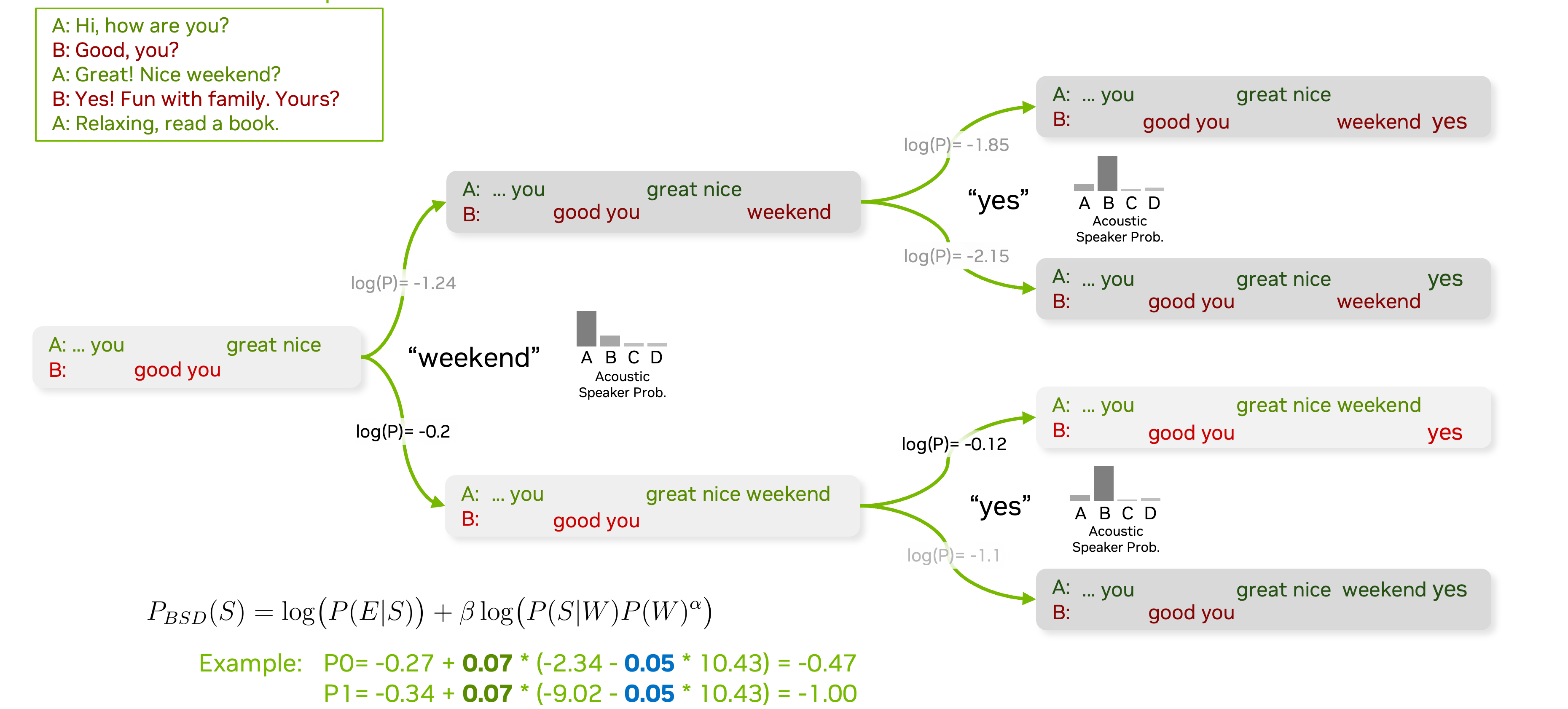}
    \caption{Example Task 2 approach based on beam-search decoding for speaker tagging \cite{park2023enhancing}}
    \label{fig:bsd_example_pic}
    \vspace{-0.0cm}
\end{figure*}


\textbf{Datasets:} The DiPCo \cite{van2019dipco}, Mixer6 \cite{brandschain2010mixer}, AMI \cite{carletta2005ami}, and CallHome American English Speech (CHAES) \cite{Canavan1997} corpora have been divided into training, development, and evaluation sets. The session names have been anonymized to prevent participants from exploiting the publicly available ground-truth data. Additionally, the evaluation scripts display the total word count and the count of erroneous words, to verify whether the output transcripts have altered the total number of words. In total, there are 222 training samples, 13 development samples, and 11 evaluation samples. The dataset is accessible through Huggingface.%
\footnote{\url{https://huggingface.co/datasets/GenSEC-LLM/SLT-Task2-Post-ASR-Speaker-Tagging}}
        


\begin{table}[tb]
\centering
\caption{Task-2 cpWER (\%) of the source files and the baseline system for text-only speaker recognition.}
\vspace{1.0ex}
\begin{tabular}{r c c}
\toprule
 System          & \textbf{dev}      & \textbf{eval}     \\ \midrule
Source Transcript    & 24.65 & 28.45 \\ \midrule
Task-2 Baseline & 24.54 & 28.37 \\ \bottomrule
\end{tabular}
\label{tab:task2_results}
\end{table}

\begin{figure}[ht]
    \centering
    \includegraphics[width=0.80\linewidth]{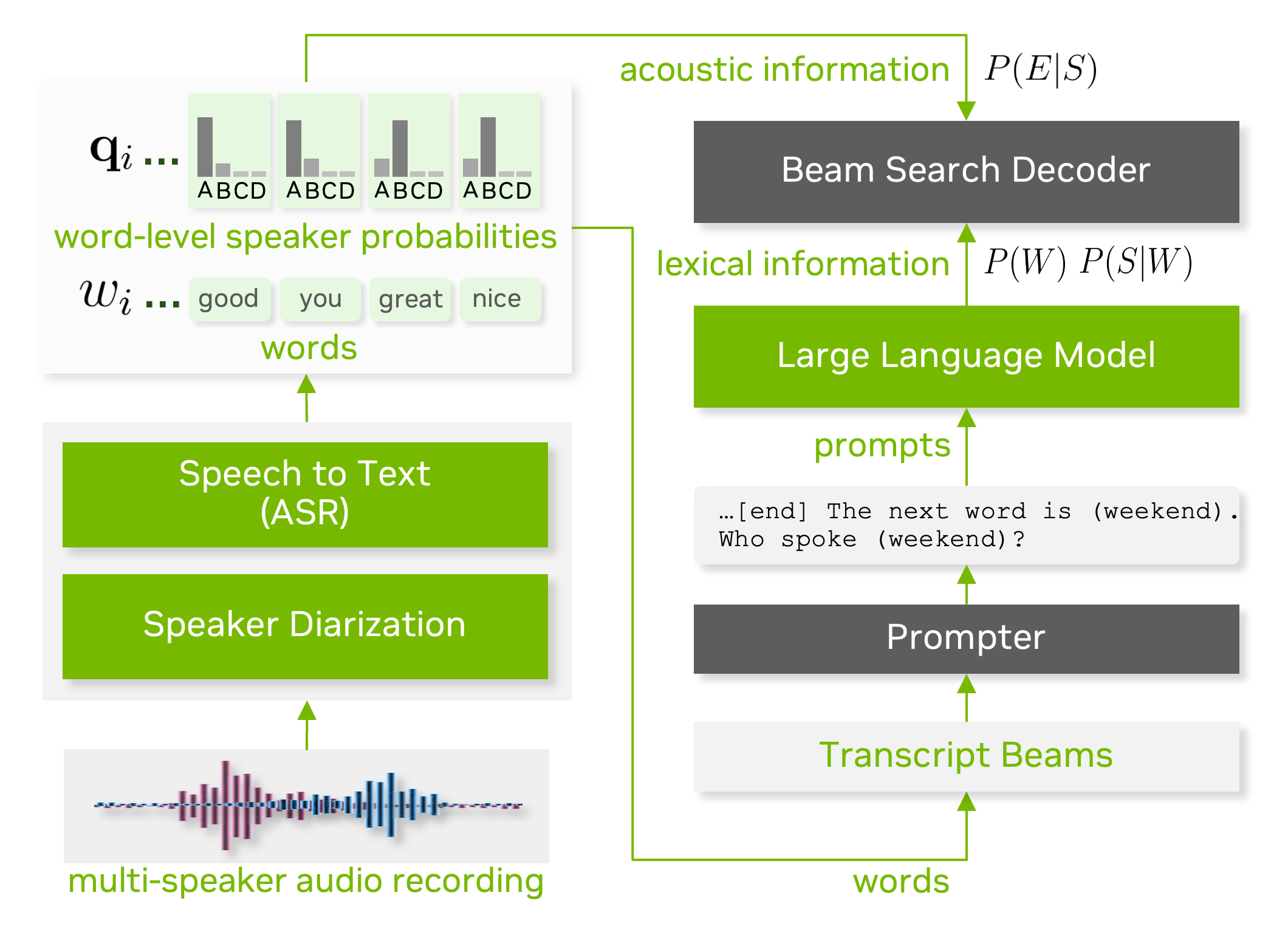}
    \caption{Dataflow for the Task 2 baseline. Note that the acoustic-only diarization probability values are set to fixed values.}
    \label{fig:bsd_pipeline}
\end{figure}

\textbf{Baseline:} 
We generated the speaker-annotated transcripts from the system proposed in \cite{park2023chime} as a baseline system,\footnote{\url{https://github.com/tango4j/llm_speaker_tagging}} based on  NeMo \cite{kuchaiev2019nemo} speaker diarization and NeMo ASR models. We provide the postprocessing method proposed in~\cite{park2023enhancing}, where we replace the LLM with n-gram language models. Since the n-gram-based system shows similar results and has low computational demands, we use this n-gram baseline to gauge the performance of beam-search-based speaker tag correction. Fig.~\ref{fig:bsd_example_pic} shows how beam search decoding can correct the speaker tagging. To mask out acoustic information, the speaker probability values in Fig.~\ref{fig:bsd_pipeline} are all fixed at 0.96. Table~\ref{tab:task2_results} shows the accuracy of the baseline for development and evaluation sets.
    
\textbf{Evaluation:} We employ concatenated minimum permutation word error rate (cpWER), as presented in \cite{watanabe2020chime}. cpWER is calculated by concatenating the speaker-wise transcripts for every label permutation and selecting the permutation that results in the lowest WER, using the open-source and publicly available MeetEval~\cite{meeteval2024} multi-speaker ASR evaluation toolkit. Additionally, we provide a Hugging Face-style leaderboard\footnote{\url{https://huggingface.co/spaces/GenSEC-LLM/task2_speaker_tagging_leaderboard}} for challenge participants to upload and evaluate their submissions.


\subsection{Task 3: Post-ASR Speech Emotion Recognition}

\textbf{Background:} Text-based SER has advanced significantly over the past decade. However, its use in real-world applications remains rare. One reason is that the majority of SER research relies on human annotation, i.e., manual transcripts. In contrast, even for elicited emotion corpora, transcripts from a state-of-the-art ASR system can result in high WERs \cite{li2023asr}, meaning that few findings obtained in the lab can be replicated in the wild. Moreover, SER on ASR transcripts is an understudied topic. Traditionally, researchers have considered confidence scores of recognized words \cite{santoso2021speech}, ASR error correction \cite{li2024crossmodal,he2024mf}, as well as fusion with audio information \cite{ang2002prosody,li2022fusing} to mitigate the side effects of ASR errors. Still, there is a lack of comprehensive studies covering diverse situations (i.e., corpora, metrics, WERs, fusion techniques). With the rise of LLMs, it has become feasible to perform SER on ASR transcripts with simple prompting \cite{feng2024foundation}. This emerging approach, however, has not been established as a reliable solution given the uneven performance with different prompting templates and the general lack of explainability of LLM outputs.


\textbf{Motivation:} Using text input only, without access to acoustic information, is a good starting point for exploring LLMs for SER, especially given that most LLMs are text-based. Insights gained about text-based SER, including handling of ASR errors, can be a foundation for future evaluations incorporating acoustic features. Potential future tasks could include ASR-error-robust multimodal fusion, enhancing word embeddings from ASR transcripts with discrete speech units, and developing ASR-integrated multimodal LLMs based on spoken language.



\textbf{Dataset:} We use the public IEMOCAP dataset \cite{busso2008iemocap}. Speech transcripts from eleven ASR models (Wav2vec2, HuBERT, Whisper, etc.) are provided for each audio segment \cite{li2024speech}. We ask participants to predict four emotion classes: angry, happy (combined with excited), neutral, and sad, for each segment. All segments are presented in the order of the conversation based on the timestamp. Participants are encouraged to use conversational context to improve emotion prediction.

\begin{figure}
    \centering
    \includegraphics[width=0.60\linewidth]{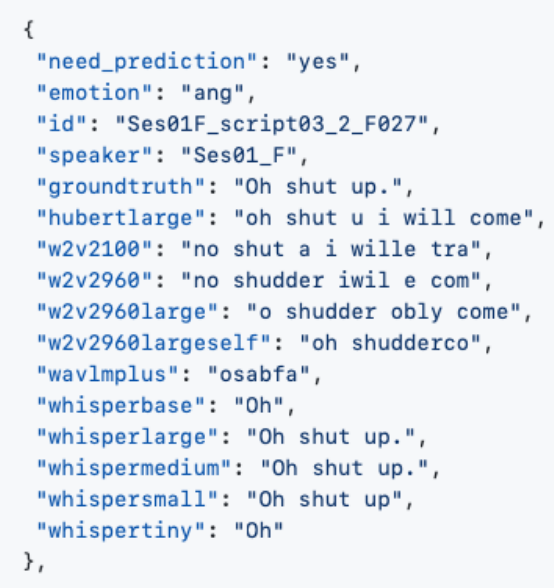}
    \caption{Example of a data entry of Task 3.}
    \label{fig:task3data}
    \vspace{-0.0cm}
\end{figure}

An exemplary data entry is shown in Fig.~\ref{fig:task3data}, where \textit{need\_prediction} indicates whether this utterance should be included in the prediction procedure. ``yes'' denotes the utterances labeled with the four emotion classes and ``no'' denotes all other utterances. Note that we have removed the utterances that have no human annotations. The key \textit{emotion} indicates the emotion label of the utterance. The key \textit{id} indicates the utterance ID, which is also the name of the audio file in IEMOCAP dataset. The ID is exactly the same as the raw ID in IEMOCAP. The key \textit{speaker} indicates the speaker of the utterance.
The key \textit{groundtruth} indicates the original human transcription provided by IEMOCAP while the remaining ten keys indicate the ASR transcription generated by the various ASR models.

\textbf{Baseline:} We provide two performance baselines with ASR transcripts from Whisper-tiny used as the text input: one with an LLM-based approach using GPT-3.5-turbo,\footnote{Version: GPT-3.5-turbo-0125; Context window: 16,385 tokens; Training data: up to Sep 2021} the other with a traditional approach based on a deep learning model. For the GPT-3.5-turbo approach, we performed zero-shot prediction with a context window of three (only previous utterances allowed), with code available\footnote{\url{https://github.com/YuanGongND/llm_speech_emotion_challenge}} to  participants as a reference. For the deep learning-based model, a two-layer feed-forward network was trained following the standard five-fold cross-validation of IEMOCAP. The first layer encodes RoBERTa output of dimension 768 into hidden states of dimension 128, and the second further encodes it into a dimension of 16. ReLU is used as the activation function between the layers. The dataset statistics and our baseline results are given in Table~\ref{task3:statistics}.

\begin{table}[bt]
\centering
\caption{Task-3 dataset statistics and baseline unweighted accuracies (\%).}
\label{task3:statistics}
\resizebox{1.0\columnwidth}{!}{
\begin{tabular}{lcc}
\toprule
 & \textbf{Training set} & \textbf{Test set} \\ \midrule
\textbf{Number of samples (all)} & 5,525 & 4,730 \\
\textbf{Number of samples (four emotions)} & 2,577 & 2,923 \\\midrule
\textbf{Baseline accuracy (GPT3.5-turbo)} & 44.70 & 55.18 \\
\textbf{Baseline accuracy (traditional)} & 62.34 & 51.08 \\ \bottomrule
\end{tabular}}
\end{table}

For the method based on GPT3.5-turbo, the accuracy is 44.70\% on the training set and 55.18\% on the test set. This significant discrepancy is reasonable since the training set contains scripted dialogs (which may not align with emotion labels), while the test set is more spontaneous. For the traditional deep learning approach, the accuracy is 62.34\% on the training set and 51.08\% on the test set. This discrepancy is also plausible, considering duplicate textual scripts in the training set, which results in overlap between the training and development subsets of the training set.


\textbf{Evaluation:} We use unweighted four-class accuracy (number of correctly predicted samples / total number of samples). We will release a training set and a test set. Participants can use the training set to develop their methods and tune hyperparameters. The test set does not come with emotion labels or ground-truth transcription and is strictly disallowed for use in model development. We will rank the models based on their accuracy on the test set, but will also further evaluate the models with a separate unpublished test set to assess generalization. Participants are free to use any LLM (such as GPT or LLaMA) or non-LLM methods (traditional text-based emotion classifiers) based on the provided ASR transcriptions. For fairness considerations, participants should not use any audio data (including audio waveforms or acoustic features) or transcribe speech using their own ASR model. Participants are allowed to use additional training datasets, as long as they are specified and publicly available, but they must not include IEMOCAP. To encourage innovation, we do not place any other restrictions on the methods used, as long as they are automated.

\section{Conclusion}
We have created the GenSEC challenge to probe the capabilities of large language models for post-processing of ASR outputs. By standardizing the tasks, datasets and metrics, we hope to create a community of researchers that will advance the state of the art in speech processing systems by loose coupling of off-the-shelf ASR systems and techniques based on generative LMs,  such as instruction prompting, text generation and in-context learning.  We propose three tasks that go beyond speech transcription correction and include text-based speaker diarization correction and emotion recognition.
Unlike traditional models, LLMs provide the potential for improving these tasks by leveraging linguistic and world knowledge learned during pretraining, and by taking advantage of long conversational context.

\clearpage
\small
\bibliographystyle{IEEEbib}
\bibliography{ref}



\end{document}